\documentclass[pdflatex,sn-nature]{sn-jnl}
\usepackage{amsmath,amssymb,amsfonts}
\usepackage{microtype}
\usepackage{xcolor}
\usepackage{colortbl}
\definecolor{proofhighlight}{HTML}{F0EBF5}
\usepackage{booktabs,tabularx,array}
\usepackage{ragged2e}
\usepackage{graphicx,placeins}

\newcolumntype{P}[1]{>{\sffamily\RaggedRight\arraybackslash}p{#1}}
\newcommand{\researchfigure}[3]{%
\begin{figure}[!tbp]
\centering
\includegraphics[width=\linewidth]{#2}
\caption{#3}\label{#1}
\end{figure}}
\unnumbered
\begin{document}
\title[Automated Mathematical Research]{Thinking Like a Mathematician: Counterexample-Guided Bidirectional Reasoning for Mathematical Discovery}
\author[1]{\fnm{Weitao} \sur{Huang}}
\author[1]{\fnm{Renzhen} \sur{Wang}}
\author[2]{\fnm{Yichen} \sur{Wu}}
\author[1]{\fnm{Deyu} \sur{Meng}}
\affil[1]{\orgname{Xi'an Jiaotong University}, \orgaddress{\city{Xi'an}, \country{China}}}
\affil[2]{\orgname{Harvard University}, \orgaddress{\city{Boston}, \state{MA}, \country{USA}}}

\abstract{
As AI integrates into mathematical research, a major unresolved risk is the pursuit of intermediate goals that are either stronger than required by the theorem or too weak to support its conclusion.
Mathematicians handle such impasses by reasoning forward from hypotheses and backward from the goal, testing proposed claims against counterexamples, and revising the goal when a route fails. Our agent workflow mirrors this practice, running the three modes concurrently and revising intermediate goals from independently checked findings. It answers Yukich's question with a sharp almost-sure transport law on the square, proves Le Gall's circle-packing conjecture, and proves positive measure for the volumes of field-free tori near axisymmetric vessels, as Grad conjectured. On nine previously solved Erd\H{o}s problems it develops alternative proofs, extensions and conjectures, plus a half-density theorem for a Problem 5 variant. Recording why each goal changed, and what it rests on, turns automated research into reasoning that mathematicians can examine, understand and build on. This study is hopeful to provide a general paradigm for building mathematical research agents inspired by how mathematicians think and reason, and shed light on how to evaluate open‑ended automated mathematical discovery beyond merely checking final‑theorem correctness.
} 

\maketitle

Artificial intelligence (AI) is taking on progressively larger parts of mathematical research. Machine-learning models trained on mathematical data first guided mathematicians towards new conjectures~\cite{Davies2021}. Systems that couple learned models with search or formal deduction then found extremal
constructions~\cite{Romera2024,Novikov2025AlphaEvolve,Georgiev2025} and solved olympiad problems~\cite{Trinh2024,Hubert2026AlphaProof,Zhang2026TongGeometry}.
More recently, research agents that combine decomposition, literature search, reasoning and verification~\cite{Liu2026Danus,Ju2026Rethlas,Feng2026Aletheia}
have begun to resolve open problems~\cite{Ju2026Rethlas,Feng2026Aletheia,Tsoukalas2026Nexus}. In most of these advances, the system is given a statement to prove or an objective to optimize, and its task is to supply the proof or the construction. A research investigation chains many such tasks whose statements are not given in advance:
only the final theorem is fixed, and each intermediate statement must be chosen along the way. Each choice shapes which objects are developed and how the
results combine into the final argument. As AI takes on longer stretches of an investigation, these choices increasingly determine the outcome, and automated research needs a principled way to make them.

Current research agents make these choices by proposing a decomposition and revising proofs or subgoals when an attempt fails~\cite{Feng2026Aletheia,Tsoukalas2026Nexus}. These revisions are guided by feedback from verification or formal search, which reports whether an attempt
succeeded and, in formal search, can show that a subgoal is false~\cite{Tsoukalas2026Nexus}. Such feedback does not by itself explain why a subgoal failed or indicate what the theorem needs in its place. The explanation matters because different failures call for different replacements: a subgoal may demand more than the theorem requires, rest on a false connection or give too little control over the conclusion. A principled choice of intermediate goals therefore needs evidence that diagnoses a failure as well as detecting it.

Mathematicians diagnose proof dead‑ends by distinguishing which constraints are intrinsic to the target theorem itself, and which are merely artefacts introduced by the tentative proof attempt~\cite{Polya1945}. 
Three complementary modes of reasoning constitute the typical toolkit human mathematicians deploy to address this very question.

\begin{itemize}
\item 
\textbf{Forward exploration:} Working forwards from the given hypotheses unfolds all mathematical structures genuinely available under the problem’s assumptions.

\item \textbf{Backward reasoning:} Working backwards from the desired conclusion reveals what properties would be sufficient to establish the result, which lets practitioners judge whether a candidate intermediate sub‑goal imposes conditions stronger or weaker than what the theorem actually demands.

\item 
\textbf{Counterexample construction:} Generating counterexamples probes the logical links assumed by a candidate proof route; a counterexample that invalidates an intermediate lemma (instead of disproving the main theorem) pinpoints exactly which constraints ought to be discarded or revised~\cite{Lakatos1976}.
\end{itemize}
These reasoning modes continuously inform one another. When hitting an obstruction, researchers can redirect their search toward alternative sufficient conditions; a newly‑found constructive example may in turn uncover a more natural proof path. Research judgement synthesizes evidence from all three lines of inquiry to select a revised intermediate question: one that leverages available structures, advances the argument, and clarifies how to resolve the encountered obstacle (Fig.~\ref{fig:research-overview}a).

\researchfigure{fig:research-overview}{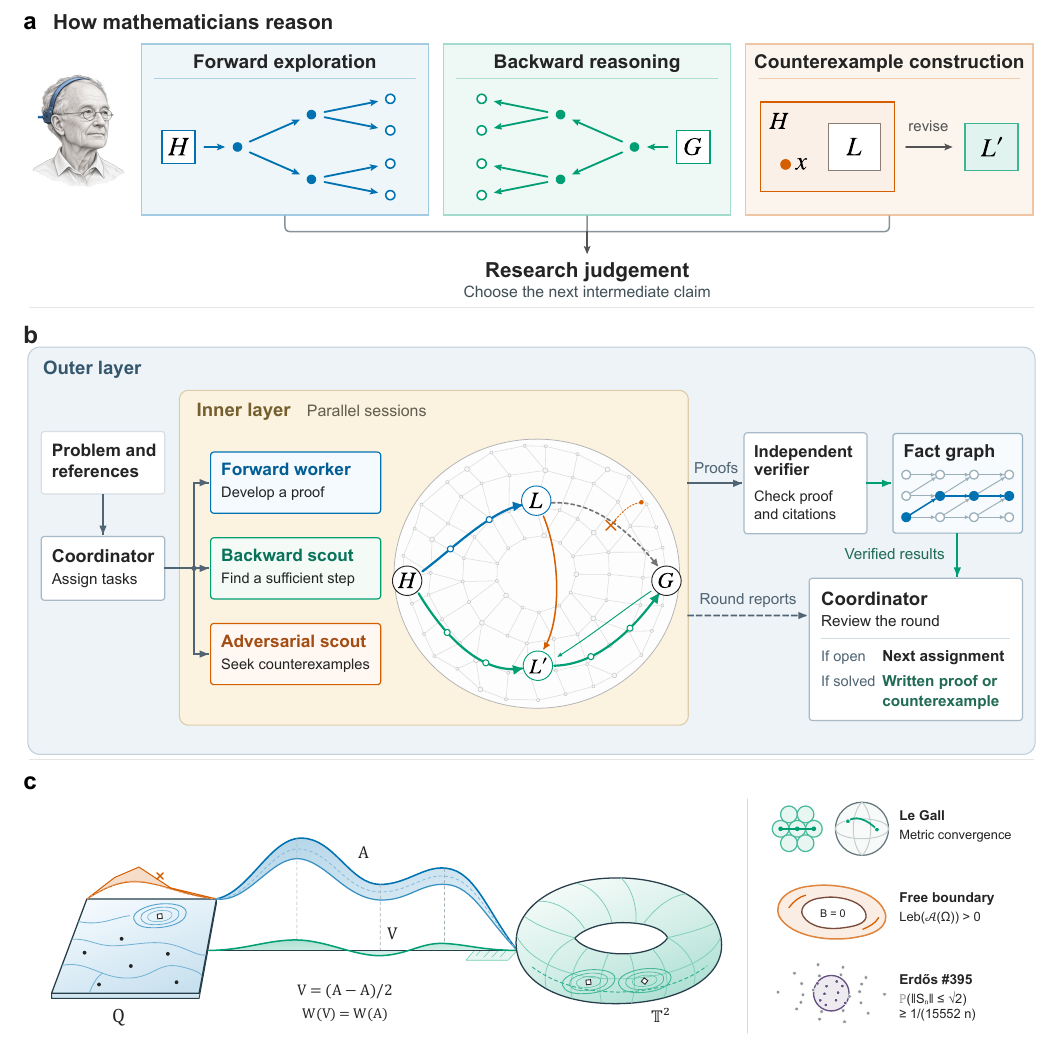}{\textbf{Choosing what to prove next.}
\textbf{a}. Mathematicians reason forward from assumptions $H$ and backward from goal $G$. A counterexample $x$ satisfies $H$ but violates $L$, prompting revision to $L'$.
\textbf{b}. Three inner agents work in parallel. The circle traces $H\to L'\to G$. In the outer layer, the verifier checks proofs and citations for the fact graph. Verified results and round reports (dashed arrow) guide the coordinator's next assignment or final answer.
\textbf{c}. In Yukich's problem, a shared boundary correction can make a profile too steep (orange). Half-differences of nearby profiles in $A$ (blue) form $V=(A-A)/2$ (green) with the same expected best Gaussian score, $W(V)=W(A)$. They can taper to zero for comparison between the square $Q$ and torus $\mathbb T^2$. At right are distance convergence at circle centres (Le Gall), equilibrium plasma volumes of positive total length in nearly axisymmetric vessels, and a probability bound for signed arrows ending near the origin (Erd\H{o}s~395).}

Based on such understanding, we organize an automated research workflow around this practice (Fig.~\ref{fig:research-overview}b). Each mode of reasoning is assigned to a separate role: 

\begin{itemize}
\item \textbf{Forward workers} develop the current task from established results.

\item \textbf{Backward scouts} reduce the target to sufficient conditions;

\item \textbf{Adversarial scouts} search for counterexamples to proposed claims. 
\end{itemize}
The roles run concurrently in independent sessions, so that each mode starts from the mathematics rather than from another role's working assumptions; the adversarial scout, for example, attacks a statement before examining its proof. Findings are exchanged as explicit mathematical objects (claims, proofs, reductions and counterexamples), and substantive results enter a shared record only after an independent verifier agent has checked them. A coordinator revises intermediate goals using the workers' reports together with checked reductions and counterexamples, which indicate why a goal should change and what should replace it. It records the reason for each change, and every accepted result carries its proof dependencies, so the
formation of research questions becomes part of an investigation that mathematicians can inspect.  

We applied the workflow to three long-standing problems. For Yukich's 1989 question~\cite{Yukich1989}, it establishes the sharp constant on the square together with almost-sure convergence along the full sample sequence. It proves Le Gall's 2014 conjecture on circle-packing embeddings of random triangulations~\cite{LeGall2014}. For Grad's 1970 free-boundary conjecture~\cite{GradNice1970}, it proves that the allowable volumes of a field-free plasma core form a set of positive Lebesgue measure in every analytic vessel sufficiently close to an axisymmetric solid torus. In each case, the recorded proof dependencies connect the final theorem to the intermediate results that followed from goal revisions. For the Le Gall and Grad problems, runs of the underlying infrastructure without the three-mode organization, conducted under different operating conditions, did not reach the targets (Fig.~\ref{fig:proof-networks}b,c). Furthermore, investigations of nine previously solved Erd\H{o}s problems develop alternative proofs, extensions and conjectures, and a variant of Problem~5 yields a uniform half-density theorem for limit points of normalized prime gaps (Table~\ref{tab:erdos}). Together, these results show that an automated workflow can choose and revise intermediate questions for explicit mathematical reasons on such problems, and can leave a record from which mathematicians can follow how each result was reached. This study thus offers a promising path toward building automated mathematical research agents that emulate how mathematicians think and reason, and provides a new and general perspective on evaluating open-ended mathematical discovery.

Typically, to intuitively illustrate how this framework operates in practice, we can consider Yukich's random‑transport problem, which shows how this organization reshapes the course of a proof
(Fig.~\ref{fig:research-overview}c).
The target is an almost-sure law, with a sharp constant, for the transport cost between $n$ uniform random points and the uniform measure on the unit square; the constant is known in the periodic setting~\cite{Yukich1989, AGGT2026}. Forward workers first pursued a natural route: correct each admissible test function near the boundary so that it becomes periodic, at arbitrarily small cost to its Lipschitz bound. The adversarial scout examined the requirement behind this route, that one correction serve every function with the same boundary values, and constructed a linear boundary trace for which any such correction loses at least the slope of the trace. Meanwhile, the backward scout's reduction of the target showed that the theorem needs only the leading coefficient, a quantity determined by the whole function family rather than by individual functions. Together, the two findings told the coordinating agent why the route failed and what the theorem actually required. It replaced the task with a comparison of localized function classes that preserves the coefficient and recorded the counterexample as the reason; forward workers then proved this comparison, which became a central step of the proof.

\section{Results}

\FloatBarrier

\subsection{How counterexamples and backward search change a proof}

When proof agents reach a difficult step, they often add a condition that would make it work and spend the next round trying to prove that condition. As the chain of reasoning grows, even a successful proof can leave it unclear which step explains the result or what would carry over to another problem. We call this repeated patching of the initial route \textbf{reasoning-chain overfitting}. Our workflow regularises the reasoning chain through two checks that mathematicians use to decide which route to pursue. The backward scout works back from the desired conclusion to find a step that would suffice. The adversarial scout tests the conditions in hand against concrete examples to find where the proposed inference fails. Together, these checks guide what the workflow tries to prove next and give it a reason to change course.

We examined these decisions in three open research problems. The first was presented by Joseph E. Yukich in \emph{Optimal matching and empirical measures} (\emph{Proceedings of the American Mathematical Society}, 1989), where he asked whether the minimum transport cost in two-dimensional random matching~\cite{AjtaiKomlosTusnady1984}, divided by its known scale, converges almost surely to the same positive constant as random points are added to a square~\cite{Yukich1989}. Scatter $n$ points independently and uniformly over a unit square, divide a fixed amount of sand equally among them and spread the sand evenly over the square at the lowest possible cost. In 1989, Yukich established that this cost falls on the scale $\sqrt{\log n/n}$. He asked whether, as points are added to the same sample, dividing the cost by this scale gives a value that settles at the same positive number~\cite{Yukich1989}. In environmental monitoring, this law could help assess how far adding measurement sites would reduce the effect of uneven coverage on an estimate of regional mean temperature.

\begin{figure}[!tbp]
{\centering
\includegraphics[width=\linewidth,height=\textheight,keepaspectratio]{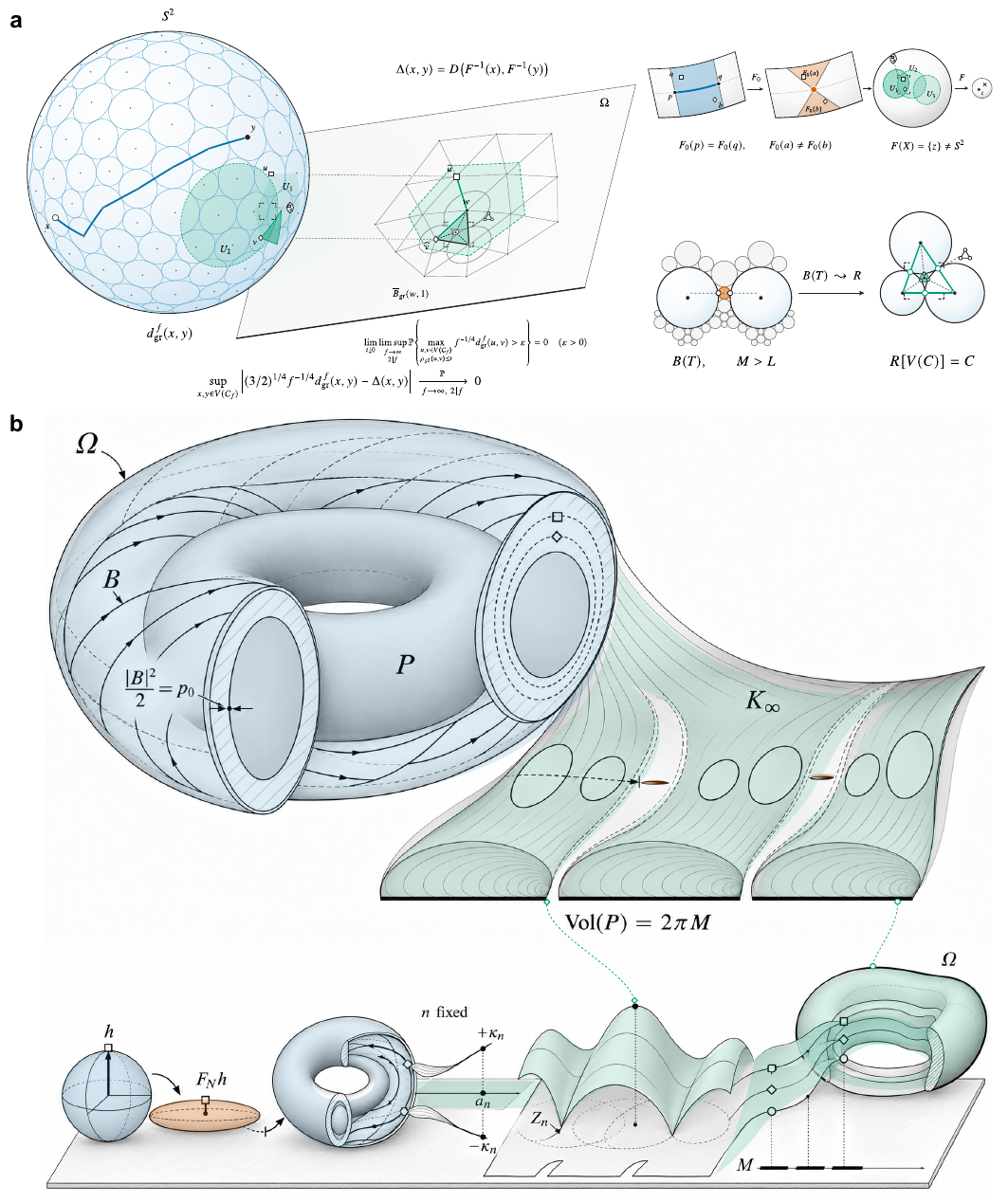}\par}
\caption{\textbf{Geometric obstructions in circle packing and vacuum equilibrium.}
\textbf{a}. The blue path illustrates graph distance between packing centres $x$ and $y$. Below, a limit map can collapse an arc, and full subdivision can distort the original packing. A refinement that retains the original tangencies, together with estimates in overlapping charts, gives uniform convergence in probability of the rescaled graph distance to the metric $\Delta$ over all pairs of original packing centres.
\textbf{b}. The vacuum field $\mathbf B$ surrounds the field-free core $P$ and balances its pressure $p_0$ at the inner wall. A near-singular direction gives a small pressure response $F_Nh$ to a unit perturbation $h$. For each nonzero angular mode $n$, invertibility at a reference point $a_n$ and analytic dependence give a zero-area singular set $Z_n$. Equilibria indexed by the retained set $K_\infty$ persist in any prescribed analytic vessel $\Omega$ sufficiently close to the axisymmetric reference. Their volumes $2\pi M$ form a set of positive Lebesgue measure.}\label{fig:proof-geometry}
\end{figure}

The second open question was conjectured by Jean-Fran\c{c}ois Le Gall in his plenary lecture \emph{Random geometry on the sphere} at the 2014 International Congress of Mathematicians in Seoul. It is that for every pair of vertices in a random triangular network, the limit of their rescaled shortest distance can be read at the original positions of their circle centres on the sphere~\cite{LeGall2014}. Draw a random network of triangles on a sphere, with a disk centred at each junction so that the disks at neighbouring junctions touch (Fig.~\ref{fig:proof-geometry}a). As the number of triangles grows, a random surface known as the Brownian sphere describes the shortest distances along the network after rescaling~\cite{Addario2017}. In his plenary lecture at the 2014 International Congress of Mathematicians, Le Gall asked whether those distances could be read at the original positions of the disk centres~\cite{LeGall2014}. In computer simulations of two-dimensional quantum gravity, this would allow distances measured on a random network to be compared directly with the distances predicted for the continuous surface~\cite{Miller2021}.

The third one was raised by Harold Grad at the 1970 International Congress of Mathematicians in Nice, where he conjectured in his lecture \emph{Mathematical problems arising in plasma physics} that the volumes a plasma can occupy in magnetic equilibrium inside a fixed three-dimensional ring-shaped vessel form a set of positive length on the number line~\cite{GradNice1970}. Consider a ring-shaped vessel whose outer wall is already built. The plasma inside contains no magnetic field and exerts outward pressure. A magnetic field in the surrounding vacuum runs along the plasma boundary and exerts the pressure that holds it in place (Fig.~\ref{fig:proof-geometry}b). From Temam's work on the shapes of confined plasma in equilibrium~\cite{Temam1975} to Kaiser and Salat's study of these boundaries~\cite{KaiserSalat1994}, a recurring question is how much freedom remains to choose the plasma shape once the vessel is fixed. We study that freedom through volume. If every volume that can be held in equilibrium is marked on a number line, do the marks together occupy a positive length? For magnetic confinement devices such as tokamaks, this idealised model could inform equilibrium design by clarifying how the shape of the vacuum vessel limits the space that the plasma can occupy.

\subsection{Yukich's limit question for transport costs on the square}

The shape of the square suggested a promising first route. Cut the square into quarters and each piece is a smaller copy of the original problem. One might therefore solve the four smaller problems and combine their answers. Some quarters, however, contain more sand than their share and others less. Even after the sand has been spread evenly within each quarter, some must still cross the dividing lines. The counterexample search tested what even a good estimate of this extra cost could establish. Suppose that the normalised costs for $n$ and $4n$ points become arbitrarily close. A sequence can satisfy that comparison and still move repeatedly between two different levels. If the swings become slow enough, the comparison between a sample size and four times that size gives the appearance of stability. This example led the workflow to look for a comparison that could pin down the value itself. Joining each pair of opposite edges of the square gives a surface for which the constant is already known~\cite{AGGT2026}.

How much can joining the edges change the cost? The proof used an equivalent way of keeping track of the work. Assign a height to each location on the square, with the difference in height between any two locations at most their straight-line distance. If a portion of sand drops one metre in assigned height, it must have travelled at least one metre. The map's score is the total drop in height, weighted by the amount of sand moved, with rises counting as negative drops. The best score over all allowed maps equals the lowest cost of moving the sand. Once the edges are joined, the heights at points that meet along a seam must agree. The workflow first tried subtracting a common correction from maps that had the same heights along their edges. Work on this repair had stalled when the adversarial scout was brought in. Its counterexample explained why. If the right edge is higher than the left by a fixed amount, the correction must rise somewhere. A map with exactly those edge heights can already slope down as steeply as allowed at that very spot. Subtracting the correction makes the slope steeper still. Changing the shape of the correction leaves it vulnerable to another such map.

The counterexample raised a question. \textbf{To recover the constant, did the proof need to preserve every height map, or only the number calculated from the maps?} For each pattern of random variation in the model, the proof takes the best score among the maps and averages these best scores across patterns. The workflow therefore grouped maps of similar shape. Within each group it took any two maps, subtracted their heights point by point and divided the result by two. Peaks and valleys shared by the two maps cancel. The small variations that remain are easy to bring gradually to zero near the edges, allowing opposite edges to meet.

A numerical example shows why this change works. Suppose the highest score in one group is $7$ and the lowest is $-3$ for a given pattern of random variation. In the new group, subtracting the lowest-scoring map from the highest-scoring map and halving the difference gives a best score of $(7-(-3))/2=5$. Reversing the sign of the variation changes the best score in the original group to $3$. The best score in the new group is still $5$. The two patterns are equally likely, so the best score in the original group also averages to $5$. \textbf{Removing the shape common to the maps preserves the group's average best score exactly.} The same argument applies wherever the score depends linearly on the map and a random pattern has the same distribution as its sign reversal. \textbf{More generally, identifying the property that determines the answer makes it possible to change the objects used in the proof while preserving that property. The same question can then be easier to answer.}

The inset in Fig.~\ref{fig:proof-networks}a shows how two parts of the investigation come together. Record~1 subjects the square and the surface with joined edges to the same random variation. It shows that the difference in their average best scores is too small to change the leading coefficient. Record~2 uses this comparison to transfer the known coefficient to the square. The proof then passes from the random model to the mean cost for the sampled points. Record~3 supplies an argument from the early exploration. Adding a point retains every point already sampled, linking the costs at successive sizes. This carries the result for the mean over to the entire sequence of growing samples, including every sample size. After division by $\sqrt{\log n/n}$, the cost approaches the same positive number $c_*=\kappa_1/(2\sqrt{\pi})$ both on average and along almost every individual sampling sequence. Here $\kappa_1$ comes from the model with joined edges~\cite{AGGT2026}. Independent large language model sessions reviewed the written proof. The Lean proof checker verified the result for the entire sequence in both Yukich's original statement and its formulation in terms of the cost of moving mass.

\subsection{Le Gall's circle-packing conjecture for graph distances}

We evaluated our workflow alongside Danus, an established agent system~\cite{Liu2026Danus}, on Le Gall's problem. Both identified the same difficulty. As the triangular network grows finer, could two junctions be drawn almost on top of one another while the shortest route between them along the network remains long, even after rescaling? Danus proposed counting such pairs. If their average number across networks could be made arbitrarily small, the chance of finding even one would also become arbitrarily small. The available results gave counts of junction pairs by the distance between them along the network. They gave no positions for the two disk centres on the sphere. Danus had identified what to count, but still needed a way to relate the route length for a particular pair to where that same pair was drawn.

Our backward scout started from the requirement that the conclusion hold for \emph{every pair}. How badly could a drawing go wrong if randomly sampled junctions always appeared to stay apart? The scout constructed a sequence of drawings of the sphere in which three landmarks stayed fixed while a whole arc was squeezed shorter and shorter. In the limit its endpoints met. The arc occupies zero area on the sphere, so a finite sample of points drawn uniformly by area will miss it with probability one. This example changed the task. The proof needed to cover a whole neighbourhood around every point, so that even a thin line that would escape random checks lay inside a region where the distances could be tested.

How could distances be checked in those regions? Danus also considered a reference drawing in which distances were already well understood. Transferring that result would require showing that every junction in the reference drawing lay close to its original disk centre. This comparison remained to be proved. Our adversarial scout tested one concrete way to relate two drawings. It subdivided all the edges and triangles and then read the positions of the old junctions from the finer network. On a network with alternating stripes, rearranging the disks after subdivision stretched the spacing along the stripes and across them by different factors. The distortion persisted however fine the network became. In parallel, a forward worker constructed a finer network that kept one original edge intact between each pair of previously adjacent junctions. When the added disks are removed, all the original contacts are still present, so the remaining disks give a drawing of the original network. The retained edges in Fig.~\ref{fig:proof-geometry}a are what allow distances studied on the fine network to be checked at the centres specified in the original question.

Using those same centres, the proof covered the network with overlapping patches to rule out the collapse seen in the arc example. Within the limiting drawing of any one patch, it showed that if two distinct positions merged, the whole patch had to shrink to a single point. Since neighbouring patches share a region, the next patch would have to shrink to the same point. Continuing across the overlaps would reduce the entire drawing to one point, contradicting the earlier result that the centres eventually reach every region of the sphere. Each patch therefore keeps its points apart. The patches fit together in a single layer on the sphere, so distant positions in different patches cannot fall on top of one another either. The kind of exceptional pair identified by the backward scout could thus be ruled out by an argument that applies to every junction at once.

\begin{figure}[!tbp]
{\centering
\includegraphics[width=\linewidth,height=0.79\textheight,keepaspectratio]{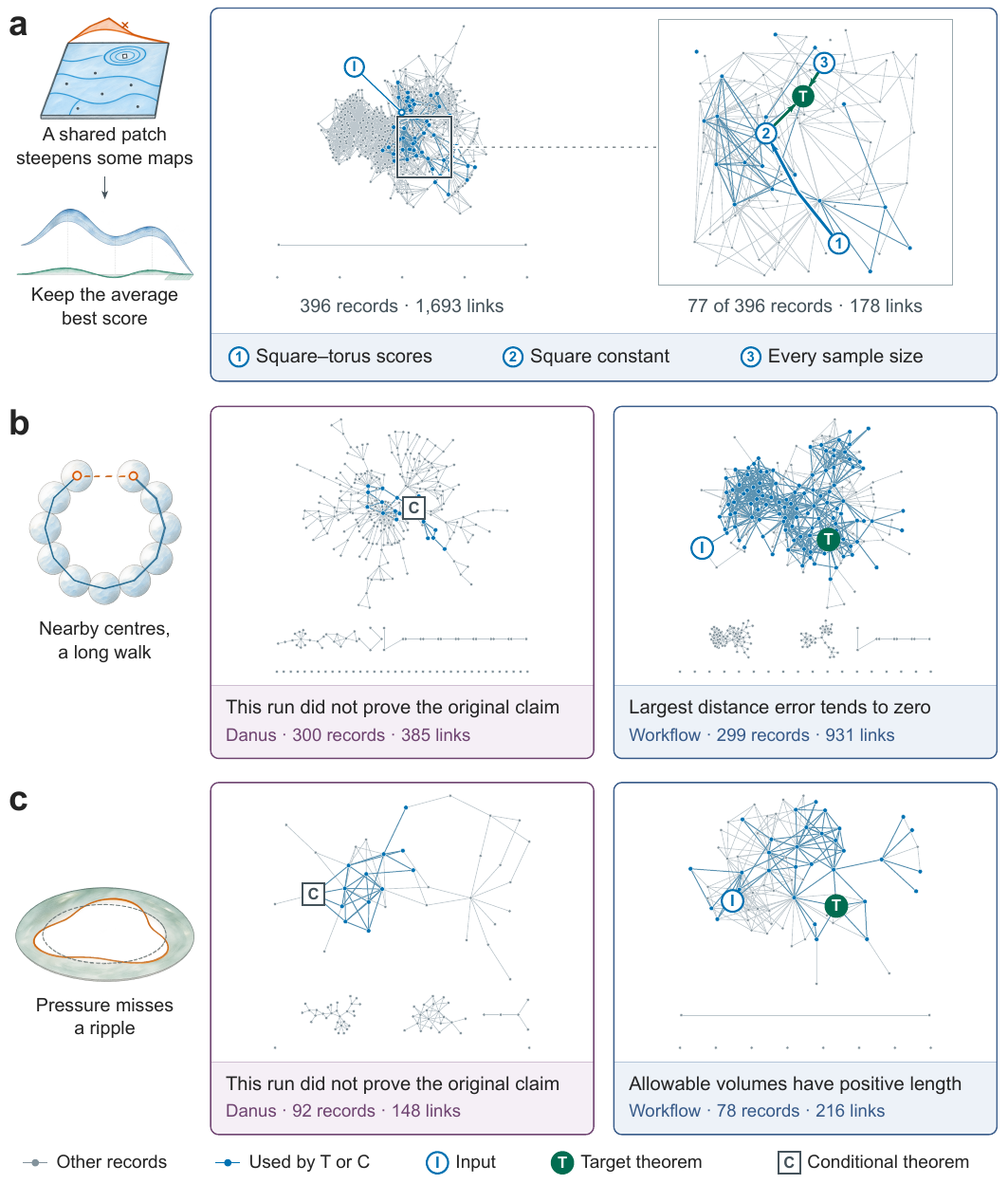}\par}
\caption{\textbf{Recorded proof dependencies in three investigations.}
Nodes are saved results. Blue traces the dependencies of a target ($T$) or conditional theorem ($C$). $I$ marks a representative input.
\textbf{a}. After a shared patch fails in Yukich's problem, comparing average best scores on the square and torus (1) gives the square constant (2). Record~3 extends the mean limit to the full sample sequence. The 77-record inset traces $1\to2\to T$ and $3\to T$, yielding the almost-sure limit $\kappa_1/(2\sqrt{\pi})$.
\textbf{b}. In Le Gall's problem the largest distance error at the original centres tends to zero in probability. Danus assumes centre coverage and agreement of distances across scales.
\textbf{c}. A ripple can escape the pressure test to first order. For each vessel within the stated analytic bounds of the rotationally symmetric reference, the allowable core volumes have positive length (33 records, 81 links). Danus obtained an interval assuming a continuous family of balanced surfaces. Both Danus runs left the original claim unresolved.}\label{fig:proof-networks}
\end{figure}

Fig.~\ref{fig:proof-networks}b compares the two routes. Danus obtained a theorem conditional on the centres spreading throughout the sphere and on drawings of different sizes giving similar route lengths at nearby positions. Our workflow proved that the drawings at different sizes approach a single continuous rule for distance at the original disk centres. The rule itself is random. As the number of triangles grows, the probability that any part of the sphere lies farther than a given positive distance from the nearest centre tends to zero. The same holds for the distance readings. Compare the distance given by this rule for two centres with the rescaled shortest route along the original network. Even for the pair with the largest error, the probability that the error exceeds any fixed positive tolerance tends to zero.

\subsection{Grad's free-boundary conjecture on allowable plasma volumes}

A core question underlying Grad’s free-boundary conjecture is whether plasma pressure can faithfully report deformations of the confining inner wall. Using a pressure difference to bring the inner wall back into balance requires knowing whether a change in pressure reveals how the wall has moved. Danus first tested different ripples in a straight tube whose ends were treated as joined. It found magnetic field settings at which the pressure failed to register a particular ripple, as well as settings at which every ripple produced a pressure change. In a curved vessel, the inner and outer sides of the bend have different shapes, so the pressure responses calculated in the straight tube still had to be established for the prescribed vessel. Our adversarial scout constructed an equilibrium directly in a curved vessel. The checks detected every change that moved the inner wall in the same way all the way around the ring. Yet when one part of the inner wall bulged and another dipped, the initial rates of change of both pressure and volume were zero. Even after all the symmetric tests had passed, the readings gave no indication of which way to correct this ripple (Fig.~\ref{fig:proof-geometry}b).

To find possible core volumes inside a fixed vessel, Danus also followed the magnetic field to identify a series of nested surfaces as candidate inner walls. A field can run along an entire surface and still be stronger in some places than in others. Danus therefore allowed a separate field adjustment for each surface and worked out when the magnetic pressure would balance a uniform internal pressure everywhere on it. It proved that if a continuous succession of surfaces met this condition, their core volumes would fill an interval. The missing step was to find such a succession in the given vessel (Fig.~\ref{fig:proof-networks}c). Our backward scout worked back from the requirement of \emph{positive length}. The allowable volumes could have gaps and still occupy a positive length in total. The task could therefore be narrowed to finding enough inner walls that could be brought back into balance, allowing settings that missed a ripple to be skipped.

How could many settings be shown to give genuinely different volumes? Danus constructed an example in which every parameter value in a short interval was assigned exactly the same inner wall. That wall was close to the shape expected at each setting, yet all the labels described the same core and gave just one point on the volume axis. To prevent this repeated counting, our backward scout proposed specifying the actual core volume at the outset and then looking for a field that could support it. The forward worker treated volume and the way the field winds around the ring as two quantities that could be adjusted separately. Starting from inner walls already in balance in the reference vessel, the worker found, for each type of ripple, a setting at which that type could be detected.

The different types of ripple could pass their tests at different settings. In this family, the pressure response near any setting is determined by its value and its successive derivatives there. The forward worker used this property to show that if a type of ripple could be detected at even one setting, the settings at which that type could go undetected together occupied zero area in the diagram whose axes record volume and field winding. The proof removed these exceptions for each type of ripple and controlled the loss of settings as progressively finer ripples were taken into account. Even when the wall was changed slightly so that it lost its rotational symmetry, a set of settings with positive total area could still be brought into balance. Volume had been one of the coordinates from the outset, so reading those settings on the volume axis gave the desired result. For every prescribed vessel sufficiently close to a rotationally symmetric reference, under the specified bounds on the wall shape and its derivatives at all orders, the core volumes that external magnetic pressure can support have a well-defined, strictly positive total length (Fig.~\ref{fig:proof-geometry}b, right, and Fig.~\ref{fig:proof-networks}c).

\subsection{Erd\H{o}s problems with new proofs, extensions and conjectures}

The three open problems showed how the workflow could select a proof route and identify what made it work. Problems with established answers offer a further test of its ability to discover new mathematics. If two tasks in an existing argument can be handled separately, the result may have a simpler proof. If the workflow can identify where a condition is used, it can also test how far the result extends. We selected ten investigations from the public collection of problems posed by Erd\H{o}s and the work that followed them. Nine began with questions that had already been answered. The tenth began with the still-open Problem~5 and studied a specific weaker version of it. Table~\ref{tab:erdos} summarises the proofs, the extensions that were established and the conjectures that remain open.

\textbf{To find a simpler proof of a known result, the workflow looks for a counterexample to a promising shortcut and works backwards from the goal to identify questions it can answer separately.} A new proof can begin by recognising that a proposed route counts the wrong objects. In Problem~395, place $n$ unit-length arrows in any directions in the plane. Toss a fair coin independently for each arrow to decide whether to follow it forwards or backwards. Beck proved that the probability of finishing on or inside the circle of radius $\sqrt{2}$ centred at the origin is at least $c/n$ for a fixed positive constant $c$~\cite{Beck1983}. The five arrows in the upper left of Fig.~\ref{fig:erdos395} give a small example. Of the 32 possible walks, 10 finish in the disk. Finding one still leaves the question of how many different walks finish there. The adversarial scout tested a natural counting strategy on a row of arrows all pointing the same way. Starting from any walk, a rule uses a fixed order of priority to flip directions and move the endpoint closer to the origin. With $2m$ identical arrows, if the last $m$ initially point forwards, all $2^m$ choices for the first $m$ lead to the same final sequence of signs. Recording many paths through these flips can therefore amount to finding a single walk that ends in the disk.

\begin{table}[!tbp]
\caption{\textbf{Proof ideas and outcomes across ten Erd\H{o}s investigations.} Nine projects revisit solved questions. The starred row gives a half-density result for Problem~5, whose full conjecture remains open. Shading highlights alternative proofs or new proof steps. Supplementary Note~7 gives the assumptions, proof records and scope of formal verification.}\label{tab:erdos}
\centering
\tablebodyfont
\setlength{\tabcolsep}{4pt}
\setlength{\aboverulesep}{1.8pt}
\setlength{\belowrulesep}{2.2pt}
\begin{tabular}{@{}P{0.046\textwidth}P{0.309\textwidth}P{0.287\textwidth}P{\dimexpr0.358\textwidth-6\tabcolsep\relax}@{}}
\textbf{ID} & \textbf{Question Summary} & \textbf{Proof Technique} & \textbf{Developments}\\
\midrule[0.5pt]
1 & How small can the largest of $n$ positive integers be if all subset sums must be distinct? & \cellcolor{proofhighlight}Chose layer sizes in a known construction~\cite{ReflectedLayers2026} by counting walks that return to their start. & The largest element grows like $C\,2^n/n^{1/3}$ with $C<3/4$ in this construction.\\
\midrule[0.25pt]
4 & Can gaps between consecutive primes reach the Erd\H{o}s--Rankin scale arbitrarily far along the number line? & Protected sites allowed by small primes while sieving the surrounding interval. & Equally long intervals can retain just those sites in the Banks--Ford--Tao model, a prediction still open for actual primes.\\
\midrule[0.25pt]
6 & Do three consecutive prime gaps satisfy $d_n<d_{n+1}<d_{n+2}$ infinitely often? & Spaced candidate primes farther apart and ruled out primes in between. & Below $X$, a run can contain at least $c\log\log\log X$ successively wider gaps.\\
\midrule[0.25pt]
29 & Can every non-negative integer be a sum of two members of $A$, with only $o(n^\varepsilon)$ ways to make $n$ for every $\varepsilon>0$? & Counted how many ways digit carries can produce the same sum in a known construction. & In certain number systems, new pairs can supply missing sums while keeping a chosen starting set intact.\\
\midrule[0.25pt]
116 & For monic $P$ with roots in the unit disk, how small is the area where $|P(z)|<1$? One case is a product of cyclotomic polynomials. & Used reflection to bound the area by measuring where a related function is negative~\cite{NPS2005}. & For these products of degree at most $d$, the smallest area shrinks like $1/\log\log d$.\\
\midrule[0.25pt]
220 & For cyclic gaps $g_i$ between residues coprime to $n$, does $\sum_i g_i^2\ll n^2/\varphi(n)$ hold? & \cellcolor{proofhighlight}Counted empty intervals of each length, including rare stretches where few integers survive the sieve. & Long gaps can occur too often for one exponential bound to hold for every $n$.\\
\midrule[0.25pt]
235 & As more primes are used to sieve the integers, do gaps between survivors, scaled by their mean, approach an exponential law? & \cellcolor{proofhighlight}Slid a window along two sieves of equal density and counted the points left inside. & For the sieves studied, bounding how much this count varies gives the same limiting pattern as independently scattered points.\\
\midrule[0.25pt]
395 & For $n$ planar unit arrows with independent fair signs, is the chance of landing within $\sqrt2$ of the origin at least $c/n$? & \cellcolor{proofhighlight}Counted pairs of endpoints in the same grid cell and saved two arrows to reach the disk (Fig.~\ref{fig:erdos395}). & For every choice of the $n$ arrows, the chance of landing in the disk is at least $1/(15552n)$.\\
\midrule[0.25pt]
457 & Are there infinitely many blocks of $\lfloor\log n\rfloor$ consecutive integers that contain a multiple of every prime up to $(2+\epsilon)\log n$ for some $\epsilon>0$? & Counted prime factors at fixed positions while varying the starting point of a qualifying block. & With primes tested up to $c$ times the block length ($1<c<2$), the standardised counts approach independent bell curves.\\
\midrule[0.25pt]
5$^{*}$ & For every $\epsilon>0$, does the prime-gap limit set $L$ cover at least a fraction $1/2-\epsilon$ of $[x,x+H]$ for all sufficiently large $H$ and all $x\geq0$? & \cellcolor{proofhighlight}Used Merikoski's result that some pair among any four distinct numbers has a positive difference in $L$. & The uniform half-density bound holds, while coverage of every non-negative value remains open.\\
\bottomrule[0.5pt]
\end{tabular}
\end{table}

Working in parallel, the backward scout kept the last two arrows in reserve and asked how far from the origin they could still bring the endpoint into the disk. The angle between them determines that range. The count for the preceding arrows then only needs to establish how often a random walk comes within reach. For example, when the number of arrows is even, the proof pairs up the remaining arrows with others pointing in similar directions. The main part of the displacement can then be written as the difference between two independent random endpoints. When the endpoints fall in the same small grid cell, that difference is small. Counting these encounters, with an allowance for the slight mismatch within each pair, gives the proof its lower bound of $1/(15552n)$ for every $n\geq1$. The grid count explains why enough walks come close to the origin, and the two reserved arrows explain why they can finish within the radius specified in the question. Lean checked both the final two-arrow step (Fig.~\ref{fig:erdos395}, lower left) and the probability bound for all $n$ (lower right).

\researchfigure{fig:erdos395}{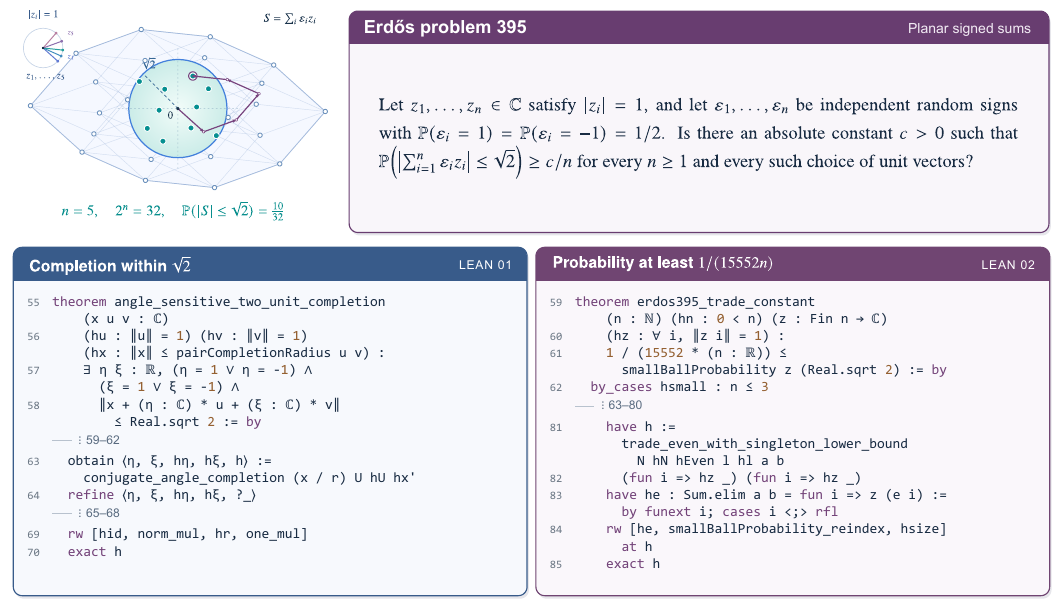}{\textbf{A two-sign completion step for Erd\H{o}s Problem~395.}
\textbf{Upper left}, Five planar unit vectors (inset) give 32 equally likely sign assignments. Teal dots mark the ten sums in the closed disk of radius $\sqrt2$ centred at the origin. Pale edges join sums whose assignments differ in one sign. The purple path adds the five signed vectors for one favourable assignment.
\textbf{Upper right}, The original problem asks for a lower bound $c/n$ on the probability of landing in the disk, uniformly over all choices of $n$ unit vectors with independent fair signs.
\textbf{Lower left}, The first Lean extract states the checked completion lemma. For unit vectors $u,v$ and a residual $x$ with $\|x\|\leq\sqrt{2+2|\det(u,v)|}$, signs $\eta,\xi\in\{-1,1\}$ can be chosen so that $x+\eta u+\xi v$ lies in the disk.
\textbf{Lower right}, The second extract states the checked probability bound $1/(15552n)$ for every $n\geq1$. Line numbers refer to the original Lean files, and dotted markers indicate omitted lines. Source Data~1 includes the full files and the build and axiom audit.}

\textbf{To find out how far a result extends, the workflow first changes the objects in the original problem and tests which parts of the explanation still hold.} In Problem~235, remove the multiples of the first few primes, one prime at a time. As more primes are included, the gaps between the remaining integers follow a known pattern. The forward worker tested whether the proportion removed at each step determines that pattern by constructing two sieves. They removed the same proportion at every step and left the same overall density of points. One left points scattered nearly at random. The other could clear an entire stretch at once. Using these examples to test the proposed extension, the adversarial scout identified a quantity that distinguishes them. Slide a window of fixed length along the number line and count the points inside. Clearing whole stretches makes that count fluctuate more. The resulting theorem applies when the screening steps are independent of one another and the statistical rule at each step is the same everywhere along the line. Suppose that even the largest proportion removed in a single step tends to zero, while the points that survive all the steps become increasingly sparse. Measure distance in units of the mean gap. It then suffices to check that, for each fixed window, the variance of its count in the limit is at most the window's length. The entire pattern approaches that of points scattered independently at an average rate of one per unit length. This single check on fluctuations replaces separate estimates of how often three, four or more positions survive together.

An existence result can also lead to a question about the objects it describes. In Problem~457, the workflow found blocks of consecutive integers arbitrarily far along the number line such that every small prime in the specified range divides at least one integer in the block~\cite{BloomErdos457}. The backward scout asked whether sharing this task affected the individual integers. If the number at one position has many prime factors, does that tell us anything about the number at another position? Consider blocks of length $\ell=\lfloor\log X\rfloor$ that cover every prime up to $c\ell$, for a fixed $1<c<2$. Choose a starting integer uniformly from the qualifying starts in $(X,2X]$. The scout proved that, as $X$ grows, the standardised counts of prime factors at any fixed selection of distinct positions each follow a bell-shaped distribution, with repeated factors included in each count. The fluctuations at those positions become independent in the limit. The existence proof thus led to a statistical description of the population of qualifying blocks.

Extending an argument to a wider class can also explain why it stops where it does. The open Problem~5 asks whether the rescaled gaps between consecutive primes come arbitrarily close, infinitely often, to every non-negative real number. Merikoski proved that among any four distinct numbers there are always two whose positive difference can be approached in this way~\cite{Merikoski2020}. The workflow asked what this fact alone implied. Consider any set of non-negative numbers whose length can be measured and that contains a positive difference from every group of four distinct numbers. The workflow proved that, for any $\varepsilon>0$, however small, the set occupies at least a fraction $1/2-\varepsilon$ of every sufficiently long interval, wherever the interval starts on the non-negative line. The scouts tested how much stronger this conclusion could be by colouring the points within a quarter of a unit of each integer. The coloured set satisfies the same four-number condition and occupies exactly half of each repeating unit, with a stretch left entirely blank. The example explains the limit of the half-coverage result. To reach every number, as the original conjecture requires, a further proof will need more information about the prime gaps themselves.

\textbf{To formulate a new conjecture, the workflow takes a choice left open by a proof and asks what further requirements that choice could satisfy.} The proof of Problem~395 found many walks with small displacement. Suppose now that each coin has a prescribed probability of heads. Could probabilities be reassigned to the walks so that every coin keeps its bias, the total displacement in each coordinate stays close to its mean on every draw, and the cost of changing the original independent distribution stays within a budget set by its fluctuations? The backward scout observed that simply discarding outcomes with large deviations can also change how often each coin comes up heads. The forward worker then tested all three requirements with two arrows pointing the same way. It chose coins with different biases for which a deviation limit of $2$ rules out two tails together. Every rare tail of the second coin must then be paired with a rare head of the first. This makes that combination much more frequent than it would be under independent tosses, exceeding the cost allowed by the conjecture. The counterexample showed that a universal deviation limit must be \textbf{strictly greater than $2$}. The workflow went on to prove that for any number of unit arrows pointing the same way and any prescribed biases, every fixed limit $2+\eta$ with $\eta>0$ suffices for all three requirements. Whether the same requirements can be met for arrows pointing in arbitrary directions is the new conjecture that arose from the proof.

\FloatBarrier
\section{Discussion}

\textbf{Deciding what to prove next is a research capability, and it should be guided by the original question rather than by the current obstacle.} A proof agent can establish a correct result in every round and still make no progress on the problem it was given. We call this reasoning-chain overfitting: a step that resists proof gives rise to further claims, even as it becomes harder to explain why the original problem requires that step. Because every new claim is checked, this failure is invisible to verification; it surfaces only when one asks what the original statement actually needs. Our workflow regularises the reasoning chain with two judgements. The adversarial scout tests whether the available assumptions support a proposed inference. The backward scout proposes an alternative task and shows how solving it would advance the original proof. Together, they give the coordinator grounds to change the forward worker's next assignment. In our investigations, these judgements led to three kinds of decision: dropping a requirement, strengthening a task and relaxing one.

In Yukich's problem, a counterexample showed which requirement to drop. The original strategy called for a single boundary patch that would work for every height map with the same boundary values. The counterexample search found a family of such maps for which any common patch that makes opposite edges agree leaves at least one map too steep. Rather than revise the patch again, we returned to the transport cost and found that the two models need to agree only on the leading term of the average best score under random fluctuations. This freed the proof to change which maps it compared. Within a group of maps of similar shape, taking half the pointwise difference of two maps cancels the features they share. Because the score is linear in the map and reversing the sign of the fluctuations leaves their distribution unchanged, the family of all such pairs preserves the group's average best score exactly (Fig.~\ref{fig:research-overview}c). The strategy is not specific to transport: when a family of objects is hard to handle one member at a time, identifying the property on which the answer depends can reveal a simpler family that retains it.

The same principle can require a proof to strengthen its next task in one problem and to relax it in another. In Le Gall's problem, the rescaled network distance between every pair of junctions must be recovered in the limit at their original disk centres. The backward scout constructed a sequence of drawings of the sphere in which the same arc shrinks until its endpoints coincide in the limit. A finite sample of locations drawn uniformly by area almost surely misses that arc, so control at typical points cannot suffice: the proof must rule out even a single pair of junctions whose centres approach each other while their rescaled network distance stays large. In Grad's free-boundary problem, by contrast, the question is whether the volumes that can be held in
equilibrium occupy a set of positive length. At some settings, the pressure has zero initial rate of change as a ripple develops on the inner wall. The proof can set those settings aside, provided that the remaining equilibria support enough distinct volumes to occupy a positive length (Fig.~\ref{fig:proof-geometry}). Neither decision could be read off from the obstacle itself. In each case, the wording of the original question determined whether the next task should cover more cases or fewer.

\textbf{Counterexamples and abandoned routes are research outputs, and a workflow should be designed to keep them.} Identifying what a proof depends on can turn a problem with a known answer into a source of new questions. In Erd\H{o}s Problem~235, two sieves left points at the same overall density, yet one could clear an entire interval at once. This difference suggested sliding a window along the number line and measuring how much the count of points inside
it varied. For the sparse sieves with independent layers studied here, consider windows whose lengths are fixed in units of the mean gap. If, for each such length, the limiting variance of the count is no greater than for independently scattered points of the same density, then the whole pattern of surviving points approaches that of independently scattered points. The answer to the original problem thus comes with a condition that can be checked directly when the argument is applied to another sieve. A counterexample can likewise sharpen a conjecture. In the extension of Problem~395, suppose that each coin must retain its probability of heads and that the cost of departing from independent tosses must stay within the stated budget. Two coins with different biases suffice to rule out a uniform bound of $2$ on the deviation of the total displacement from its mean, whereas for arrows pointing in the same direction any fixed bound above $2$ suffices. This case marks where the difficulty lies and provides a benchmark for the conjecture that some uniform bound meets the same requirements for arrows pointing in arbitrary directions.

Routes that are set aside can also supply part of a final proof, provided the workflow records them. Early work on Yukich's problem used the fact that adding a point leaves the previous sample intact, and showed how to carry a result about mean transport cost over to the cost along a single sequence of samples as points are added one by one. When the comparison between the square and the
torus later determined the constant, this earlier argument supplied the other part of the final proof (Fig.~\ref{fig:proof-networks}a). Two records made this reuse possible. The reports written at the end of each round preserve the mathematical reason a step was set aside, and the fact graph retains every reusable result together with its complete proof and the earlier results on which it depends. In a finished paper, such branches rarely survive. Read together, these records let a researcher trace which steps a counterexample affects and recover useful results from branches on which work has stopped.

\textbf{The choice of the next task can be measured, both in research agents and in the people who learn from them.} The free-boundary problem shows how two systems can diverge at the same obstacle. Danus had already shown that many parameter values could describe the same inner wall, and hence a single volume. Our backward scout instead made the actual volume part of the claim to be proved: it proposed fixing the volume first and then seeking a magnetic field that could support it, so that the construction addressed the quantity in the original question directly (Fig.~\ref{fig:proof-networks}c). Both native Danus runs (Fig.~\ref{fig:proof-networks}b,c) gave priority to accumulating checked results. Because both systems keep step-by-step records, one can compare what each established at the shared obstacle and what each still had to prove. A controlled evaluation of each scout's contribution would start different configurations from the same unfinished proof at the point where it stalled, with the model, computational budget and stopping rule held fixed. It would then ask two questions: does each configuration find counterexamples that withstand scrutiny, and do the new tasks it assigns ultimately advance the proof of the original claim?

The same records could help students practise a skill that finished proofs conceal: asking a better question after an attempt fails~\cite{Lakatos1976}. In Erd\H{o}s Problem~395 (Fig.~\ref{fig:erdos395}), students could start from several different choices of signs for a row of arrows pointing the same way and apply the same sign-flipping rule. They would see that different starting arrangements lead to the very same final arrangement, whose walk ends in the disk. A teacher could then ask what else is needed to go from one successful walk to a lower bound on the proportion of walks that succeed. A student who asks how many starting arrangements can lead to a given successful arrangement, and proposes to bound that number, has turned the counterexample into a new proof task. An AI tutor could supply a checked example when a student gets stuck, and a counting problem in a different setting could then test whether the student recognises the same gap unaided. The judgement we ask of research agents is thus one that can also be taught and assessed in people.

\textbf{As verification becomes routine, the bottleneck for AI in mathematics shifts from checking proofs to choosing questions.} The most persistent failure we observed was not an incorrect proof but a sequence of correct ones that led away from the problem. Verification certifies that a step is valid, not that it is worth taking, so an agent rewarded for checked results will produce more of them. At the same obstacle in the free-boundary problem, Danus and our workflow differed less in what they could prove than in what they chose to prove next (Fig.~\ref{fig:proof-networks}b,c). Future agents should therefore be built to ask what the original statement requires, and their counterexamples and abandoned routes should be released as results, since each is checkable and reusable. The hardest problem is evaluation: whether a new task advances the proof is often clear only once the proof is complete, so assessing research judgement means crediting good choices before their payoff is known. Mathematics is the field where this can be studied most precisely. The proofs an agent completes show what it can do; the questions it chooses show whether it is doing research.

\section{Methods}
\subsection{Workflow overview}
Existing agent systems for mathematical proof can fall into the trap of combining elaborate workflows and prompts with ever more computation to force a proof along a chosen route. When a step resists proof, an agent may add a condition and extend the argument to justify it. We call this repeated patching of the route chosen at the outset \emph{reasoning-chain overfitting}. The proof can then become a long chain of conditional arguments that is difficult to follow, making it increasingly hard to explain how the conditions of the original problem work together.

To enable agents to judge whether a route remains worth pursuing as they develop a proof, we designed a workflow driven by large language models (LLMs), such as GPT and DeepSeek, with an inner and an outer layer (Fig.~\ref{fig:research-overview}b). The inner layer retains the standard forward worker for proof development and introduces two scout roles, with all three working in parallel in separate sessions. The backward scout works back from the goal to formulate an alternative route as a concrete question, then proves that resolving the question would be sufficient to reach the goal. The adversarial scout seeks counterexamples to key steps along candidate routes to identify early where an argument fails and why.

To turn the inner layer's parallel investigations into a continuing research process, the outer layer adopts the coordinator and independent verifier roles common in multi-agent systems. The coordinator receives the researcher's problem and reference materials, keeps the original problem as the shared goal across rounds, and writes assignments specifying the question to address, where to start and what counts as completion. The verifier checks each complete proof in a separate session, retrieving cited results to check that they apply where used and returning the proof for revision when it finds a gap, with the precise step identified. To make established results available to later investigations, we include a shared fact graph that stores verified conclusions and their complete proofs, with each node representing a proved conclusion and edges connecting it to the earlier results used in its proof.

Even an investigation that only identifies where a route is blocked can help the coordinator decide what to do next, so each inner agent also prepares a report at the end of the round. The coordinator reviews the reports alongside the fact graph, decides which routes to pursue and distributes written assignments for the next round's questions to the inner agents. Once the original problem is resolved, the workflow delivers a written proof or counterexample.

\subsection{Forward worker}
As in other systems for mathematical proof, a forward worker develops proofs from established results. Which claim it is assigned to prove determines the direction of that work. If every new task is drawn from the same proof route, changing individual assignments still leaves the system following that route. In our workflow, the coordinator can use the two scouts' findings to choose another route and assign its next unresolved question to a forward worker.

At the start of each round, the worker reads its assignment to identify the claim it is to prove, then retrieves results from the fact graph that can serve as a starting point. In the Yukich project, a result establishing the torus constant was already available, and the worker was assigned to compare the square and the torus under the same random fluctuations and establish the required relation between them.

The worker submits the new conclusion and its complete proof to the verifier, citing the records used so that each step can be checked. In the Yukich project, this submission combined the verified comparison (record~1 in Fig.~\ref{fig:proof-networks}a) with the established torus constant to determine the leading constant for the square's Gaussian model (record~2). The round report states which conclusions passed verification and, for an incomplete task, identifies the last step proved and the next one to be established. The coordinator can assign this gap as a separate task for the next round or ask the adversarial scout to test the step first.

\subsection{Backward scout}
A single goal can admit several proof routes, and mathematicians often start by asking what must ultimately be established before working back to conditions that would guarantee it. If a forward worker also undertakes this task while developing a proof, its search can remain tied to the claim at hand. If the coordinator explores each alternative itself, this work competes with reading reports and assigning tasks. We therefore assign a dedicated backward scout to work from the goal and identify questions for the next round whose solutions can lead back to the original problem.

The backward scout receives the goal for the round, works back to conditions that would guarantee it, and queries the fact graph to see how far the available results carry each route. In Le Gall's circle-packing problem, it found a result showing that vertices with small rescaled graph distance also have nearby circle centres on the sphere. Where the available results leave a gap, the scout tries relaxing a requirement or omitting a step, then checks whether the remaining conditions still suffice to reach the original goal.

Once the scout can show that a condition suffices to reach the goal, it submits this argument separately to the verifier. In its round report, it cites the accepted record and formulates the condition still to be proved as a question ready to be assigned. In round~36 of the Le Gall project, the report turned the missing reverse comparison into a concrete task of ruling out even a single pair of vertices whose circle centres approach each other on the sphere while their rescaled graph distance remains large. The coordinator can then identify the next step to assign to a forward worker and explain how proving it connects to the original problem.

\subsection{Adversarial scout}
Even with forward and backward exploration under way, the workflow may remain blocked by an intermediate claim that resists both proof and refutation. To test whether the step holds, the adversarial scout reads the claim and its conditions from the assignment, then tries to construct as simple an example as possible in which all the conditions hold and the proposed conclusion fails. When there are many candidates to examine, it can run a script to screen them and save the results so that later searches can build on the cases already checked.

When submitting a counterexample, the scout states which conditions the example meets and identifies where the proposed claim fails. In Grad’s free-boundary volume problem, it found an example in which the existing checks detected movement when the inner wall moved inwards or outwards all the way around the ring. The same checks missed a ripple in which some parts of the wall bulged out and others receded. This example showed why local variations also needed to be checked after the tests of movement around the whole ring had passed.

After verification, the scout places the counterexample record beside the refuted claim in its report, allowing the coordinator to identify the tasks that depend on that step and choose the next task from the alternatives proposed by the backward scout. If no counterexample is found in the round, the report states which cases were examined so that the search can continue in the next round.

Some examples satisfy only part of the original problem's conditions while meeting every condition actually used in a proposed shortcut. If the conclusion fails in such an example, the scout can identify the unsupported step in that shortcut. The coordinator can then assign a more specific task to a forward worker by asking which unused condition of the original problem can rule out the example. The resulting proof can explain why the earlier argument failed and which condition enabled the new route to overcome the same obstacle.

\bibliography{nmi-references}
\end{document}